\documentclass[journal]{IEEEtran}

\usepackage[utf8]{inputenc}
\usepackage[T1]{fontenc}
\usepackage{graphicx}
\usepackage{booktabs}
\usepackage{amsmath,amssymb}
\usepackage{url}
\usepackage{hyperref}
\usepackage{xcolor}
\usepackage{multirow}
\usepackage{cite}
\usepackage{amssymb}

\hypersetup{
    colorlinks=true,
    linkcolor=black,
    citecolor=black,
    urlcolor=blue!60!black,
}

\begin{document}

\title{Zero-Shot Confidence Estimation for Small LLMs: When Supervised Baselines Aren't Worth Training}

\author{Luong N. Nguyen
\thanks{Code and data available at: \url{https://github.com/BuffaloTechRider/zero-shot-llm-confidence-estimation}}}

\markboth{Preprint, May 2026}{}

\maketitle

\begin{abstract}
How reliably can a small language model estimate its own correctness? The answer determines whether local-to-cloud routing---escalating queries a cheap local model cannot handle---can work without supervised training data. As inference costs dominate large language model (LLM) deployment budgets, routing most queries to a cheap local model while reserving expensive cloud calls for hard cases is an increasingly common cost-control strategy. We compare zero-shot confidence signals against RouteLLM-style supervised baselines across three 7--8B model families and two datasets (${\sim}$1,000 and ${\sim}$500 queries per model, respectively). Average token log-probability, which requires no training data, matches or exceeds supervised baselines in-distribution (Area Under the Receiver Operating Characteristic curve (AUROC) 0.650--0.714 vs.\ 0.644--0.676) and substantially outperforms them out-of-distribution (0.717--0.833 vs.\ 0.512--0.564), because it measures a property of the model's generation rather than the query distribution. This paper further proposes retrieval-conditional self-assessment, a pre-generation signal that selectively injects retrieved knowledge when similarity is high, improving over bare self-assessment by up to +0.069 AUROC at 3--10$\times$ lower latency than log-probability. A supervised baseline trained on 1,000 labeled examples never exceeds the zero-shot signal. We release all code, data, and experiment logs.
\end{abstract}


\section{Introduction}
\label{sec:intro}

\IEEEPARstart{K}{nowing} when an LLM, especially a small or cheap language model, will answer correctly is a prerequisite for deploying it reliably. The most immediate application is local-to-cloud routing: running a cheap local model for most queries and escalating to an expensive cloud model only when the local model is likely to fail. The routing decision depends entirely on the quality of the confidence estimate.

The dominant approach trains a supervised classifier on labeled examples to predict whether the local model will succeed~\cite{ong2024routellm}. This works when training and deployment distributions match. But real deployments face shifting query distributions, new domains, and cold-start scenarios where no labeled data exists. This paper shows that this brittleness is not incidental but structural: supervised confidence estimators learn properties of the \emph{query distribution} (which embedding regions correspond to easy questions), and those properties do not transfer.

Zero-shot confidence signals offer an alternative. Rather than learning query patterns, they measure properties of the \emph{model's own generation}---how surprised it was by its output, or how confident it reports being. These signals require no training data and are available from the first query. The question is whether they are accurate enough to replace supervision.

We find that they are. Average token log-probability, the simplest zero-shot signal, matches or exceeds supervised baselines in-distribution and substantially outperforms them when the query distribution shifts. We further propose a novel pre-generation signal, retrieval-conditional self-assessment, that improves over bare self-assessment by selectively grounding the model's confidence judgment in retrieved knowledge, at 3--10$\times$ lower latency than log-probability.

The conceptual distinction is between \emph{query-side} and \emph{generation-side} confidence signals. Supervised estimators learn a mapping from query embeddings to difficulty, a property of the input distribution. Zero-shot signals measure properties of the model's generation process, including token-level surprise, self-reported confidence, or output consistency. If the query distribution shifts, the supervised mapping breaks, but the generation-side signal should still work because it does not depend on the input distribution. This distinction has been discussed conceptually~\cite{kadavath2022language,chuang2025confident} but not tested head-to-head: no published work compares zero-shot and supervised confidence signals across multiple model families \emph{and} multiple datasets. We take a first step toward filling this gap.
\subsection{Contributions}

\begin{enumerate}
\item The finding that zero-shot log-probability transfers across both models and datasets while supervised routing collapses out-of-distribution (AUROC 0.717--0.833 vs.\ 0.512--0.564 on TriviaQA), explained by the mechanistic distinction between query-side and generation-side signals (Section~\ref{sec:cross-dataset}).
\item Retrieval-conditional self-assessment (GSA~v3), a novel pre-generation confidence signal that selectively injects retrieved knowledge when similarity is high and falls back to a prompt indistinguishable from no-retrieval otherwise, improving over bare self-assessment by up to +0.069 AUROC at 3--10$\times$ lower latency than log-probability (Section~\ref{sec:retrieval-gsa}).
\item A learning curve showing supervised routing converges at ${\sim}$250 examples but never exceeds the zero-shot baseline, even with 1,000 labeled examples (Section~\ref{sec:learning-curve}).
\item A cross-model, cross-dataset evaluation framework (3 models $\times$ 2 datasets $\times$ ${\sim}$1,000 queries each) with bootstrap confidence intervals (CIs) and paired significance tests, a cold-start deployment guide (Section~\ref{sec:deployment}), and all code, data, and experiment logs released for full reproducibility.
\end{enumerate}

All experiments are reproducible on a single laptop with Ollama and AWS Bedrock. Total cloud cost: \$123.

\section{Signals and Baselines}
\label{sec:signals}

\subsection{Zero-Shot Confidence Signals}

We evaluate four zero-shot signals computed at inference time without training data.\footnote{We also tested a keyword-based query classifier (factual/real-time/creative heuristic) and an energy scorer (logistic regression on query embeddings). Both performed at or below chance on all models and are omitted from the main results. Full numbers are in the released data.}

\textbf{Log-probability (logprob).} Average per-token log-probability of the local model's generated answer, mapped to $[0,1]$ via the sigmoid function $\sigma(x) = 1/(1+e^{-x})$ applied as $\sigma(\bar{\ell} \cdot 2 + 3)$, where $\bar{\ell}$ is the mean token log-prob and the constants center the transition around typical log-prob values ($\bar{\ell} \approx {-}1.5 \mapsto 0.5$). Higher values indicate greater token-level confidence~\cite{kadavath2022language}. Since AUROC depends only on score ranking and the sigmoid is monotonic, the specific constants do not affect our reported AUROC; they matter only when choosing a deployment threshold.

\textbf{Self-assessment (GSA).} A single-token YES/NO probe: the model is asked ``Are you confident you can answer this question correctly?'' and the confidence score is the normalized probability of YES over the two options: $P(\text{YES}) = e^{\ell_{\text{YES}}} / (e^{\ell_{\text{YES}}} + e^{\ell_{\text{NO}}})$, where $\ell_{\text{YES}}$ and $\ell_{\text{NO}}$ are the log-probabilities of the YES and NO tokens at the first generated position. Unlike logprob, no sigmoid is needed because the two-class normalization already produces a $[0,1]$ probability. We propose a retrieval-conditional variant~(v3) that injects strong retrieved hits when available but falls back to a byte-identical bare prompt otherwise, so the model cannot distinguish ``retrieval found nothing'' from ``retrieval was never attempted.'' This design prevents the model from being primed toward NO by the absence of context---a failure mode we identified empirically (Section~\ref{sec:retrieval-gsa}). We also test a bare variant~(v2) as an ablation baseline.

\textbf{Self-consistency (SC).} Token-overlap Jaccard similarity coefficient between two independent generations at temperature 0.0 and 0.7~\cite{wang2022selfconsistency}.

\textbf{Knowledge similarity (KS).} Maximum cosine similarity between the query embedding and top-5 retrieved KB entries via FAISS inner-product on L2-normalized bge-large-en-v1.5 embeddings (1024-dim).

\subsection{Supervised Baselines}

Following the embedding-based approach in RouteLLM~\cite{ong2024routellm}, we train two logistic regression classifiers (LogisticRegressionCV, 5-fold) on a disjoint 1,000-query training corpus per model. This is a simplified reimplementation: the full RouteLLM system uses preference data and more expressive architectures. Our baseline isolates the query-embedding signal to test the query-side vs.\ generation-side distinction cleanly.
\begin{itemize}
\item \textbf{RouteLLM-nm}: query embedding $\to$ $P(\text{local sufficient})$
\item \textbf{RouteLLM-pks}: $[\text{query embedding} \| \text{knowledge similarity}] \\ \to P(\text{local sufficient})$
\end{itemize}

\section{Experimental Setup}
\label{sec:setup}

\subsection{Models}

Three local models at similar scale (7--8B parameters) from different families and Reinforcement Learning with Human Feedback (RLHF) programs are shown in Table~\ref{tab:models}. Cloud model: Claude Sonnet~4.5 (AWS Bedrock). Judge: Claude Opus~4.5.

\begin{table}[t]
\centering
\caption{Local Models Evaluated}
\label{tab:models}
\begin{tabular}{lrl}
\toprule
Model & Params & RLHF Emphasis \\
\midrule
Qwen-2.5-7B & 7.6B & Calibration-aware \\
Llama-3.1-8B-Instruct & 8.0B & Helpfulness \\
Mistral-7B-Instruct & 7.2B & Mixed \\
\bottomrule
\end{tabular}
\end{table}

\subsection{Datasets}

\textbf{MMLU-Pro} (primary): 10-option MCQ, 14 categories. KB: 995 seeded entries. Eval: 931--999 queries per model.

\textbf{TriviaQA} (secondary): open-ended short-answer. KB: 200 seeded entries. Eval: 500 queries per model.

\subsection{Labeling}

MMLU-Pro uses regex letter extraction with LLM-judge fallback. TriviaQA uses standard substring matching. Judge fallback rates are shown in Table~\ref{tab:judge}. Mistral's 97\% judge rate reflects poorly formatted responses; its AUROC is more sensitive to judge noise. Label noise bound: 13\% inter-judge disagreement (Opus vs.\ Sonnet audit on 100 rows). Differential comparisons are robust to symmetric noise.

\begin{table}[t]
\centering
\caption{Judge Fallback Rates on MMLU-Pro}
\label{tab:judge}
\begin{tabular}{lrr}
\toprule
Model & Judge Calls / Total & Rate \\
\midrule
Qwen-2.5-7B & 688 / 931 & 74\% \\
Llama-3.1-8B & 755 / 997 & 76\% \\
Mistral-7B & 972 / 999 & 97\% \\
\bottomrule
\end{tabular}
\end{table}

\subsection{Evaluation Protocol}

AUROC with 1,000-sample bootstrap CIs (fixed seed). Paired deltas via bootstrap resampling on the same query set. Significance: 95\% CI excludes zero.

\section{Results}
\label{sec:results}

\subsection{Per-Signal AUROC on MMLU-Pro}

Log-probability is the best single signal on every model (Table~\ref{tab:main}). It significantly exceeds RouteLLM on Qwen ($\Delta = +0.049$, CI $[+0.008, +0.087]$) and ties on Llama and Mistral. Local accuracy: Qwen 43.7\%, Llama 36.2\%, Mistral 30.5\%.

\begin{table*}[t]
\centering
\caption{AUROC on MMLU-Pro (95\% Bootstrap CI in Brackets). Best Zero-Shot Signal in Bold. \\ Number of evaluation samples $n \approx 950$ Per Model.}
\label{tab:main}
\begin{tabular}{lccc}
\toprule
Signal & Qwen-2.5-7B & Llama-3.1-8B & Mistral-7B \\
\midrule
\textbf{logprob} & \textbf{0.714} {\scriptsize [.683,.746]} & \textbf{0.650} {\scriptsize [.616,.687]} & \textbf{0.678} {\scriptsize [.644,.716]} \\
GSA v3 & 0.562 {\scriptsize [.522,.598]} & 0.614 {\scriptsize [.577,.652]} & 0.638 {\scriptsize [.601,.674]} \\
SC & 0.504 {\scriptsize [.490,.517]} & 0.594 {\scriptsize [.558,.627]} & 0.604 {\scriptsize [.569,.638]} \\
KS & 0.426 {\scriptsize [.389,.465]} & 0.413 {\scriptsize [.379,.452]} & 0.422 {\scriptsize [.387,.459]} \\
\midrule
RouteLLM-nm & 0.664 {\scriptsize [.629,.699]} & 0.662 {\scriptsize [.628,.701]} & 0.676 {\scriptsize [.641,.710]} \\
RouteLLM-pks & 0.665 {\scriptsize [.629,.700]} & 0.644 {\scriptsize [.609,.683]} & 0.676 {\scriptsize [.641,.711]} \\
\bottomrule
\end{tabular}
\end{table*}

\subsection{Cross-Dataset Transfer}
\label{sec:cross-dataset}

\begin{figure}[t]
\centering
\includegraphics[width=\columnwidth]{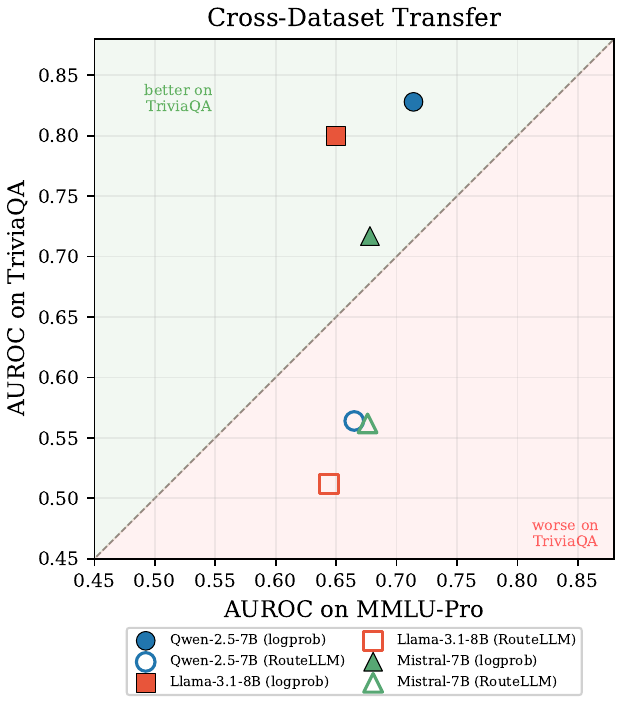}
\caption{Cross-dataset transfer. Filled markers: logprob (above diagonal = better on TriviaQA). Open markers: RouteLLM (below diagonal = collapses on TriviaQA).}
\label{fig:cross}
\end{figure}

The sharpest test of our thesis---that generation-side signals transfer while query-side signals do not---is cross-dataset evaluation. RouteLLM, trained on MMLU-Pro embeddings, collapses from 0.662 to chance (0.546) on TriviaQA (Table~\ref{tab:cross}); per-model: Qwen 0.564, Llama 0.512, Mistral 0.562. Log-probability, which measures the model's token-level surprise at its own output, \emph{improves} to 0.782 (per-model: Qwen 0.828, Llama 0.800, Mistral 0.717). Fig.~\ref{fig:cross} visualizes this divergence.

\begin{table}[t]
\centering
\caption{Cross-Dataset AUROC (Averaged Across 3 Models)}
\label{tab:cross}
\begin{tabular}{lrrr}
\toprule
Signal & MMLU-Pro & TriviaQA & $\Delta$ \\
\midrule
logprob & 0.681 & \textbf{0.782} & $+$0.101 \\
GSA & 0.605 & 0.703 & $+$0.098 \\
RouteLLM & 0.662 & 0.546 & $\mathbf{-0.116}$ \\
\bottomrule
\end{tabular}
\end{table}

\subsection{Supervised Learning Curve}
\label{sec:learning-curve}

\begin{figure}[t]
\centering
\includegraphics[width=\columnwidth]{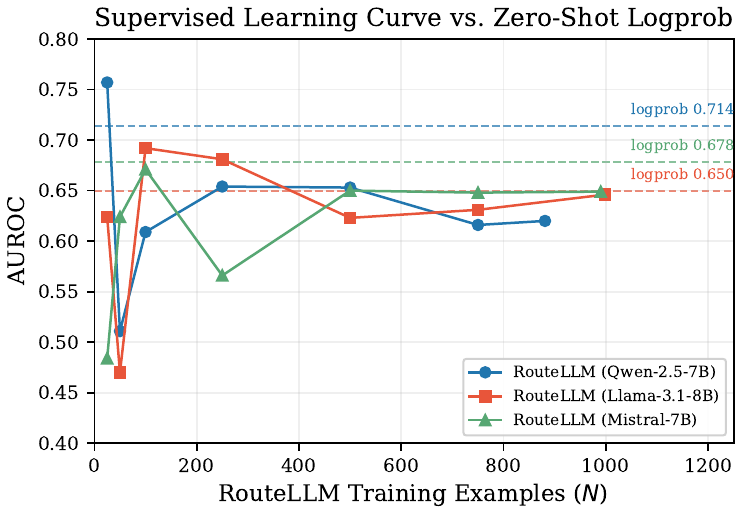}
\caption{RouteLLM learning curve (solid) vs.\ logprob zero-shot AUROC (dashed), per model.}
\label{fig:lc}
\end{figure}

RouteLLM is unstable at small $N$, converges around $N{=}250$--$500$, and at convergence never consistently exceeds the zero-shot line (Table~\ref{tab:lc}, Fig.~\ref{fig:lc}). At $N{=}25$, individual models range from 0.484 (Mistral) to 0.757 (Qwen)---a 0.27 spread reflecting high variance rather than genuine signal quality.

\begin{table}[t]
\centering
\caption{RouteLLM-nm AUROC at Increasing Training Sizes. Last Row: Logprob (Zero-Shot).}
\label{tab:lc}
\small
\begin{tabular}{rrrr}
\toprule
Training $N$ & Qwen & Llama & Mistral \\
\midrule
25 & 0.757 & 0.624 & 0.484 \\
50 & 0.511 & 0.470 & 0.624 \\
100 & 0.609 & 0.692 & 0.671 \\
250 & 0.654 & 0.681 & 0.566 \\
500 & 0.653 & 0.623 & 0.650 \\
${\sim}$1000 & 0.620 & 0.646 & 0.649 \\
\midrule
logprob (0) & \textbf{0.714} & \textbf{0.650} & \textbf{0.678} \\
\bottomrule
\end{tabular}
\end{table}

\subsection{Signal Latency}

Log-probability requires 1.5--4.8s of local generation before routing (Table~\ref{tab:latency}). A two-stage system can use GSA as a pre-generation filter (${\sim}$500ms) and logprob as a post-generation quality check.

\begin{table}[t]
\centering
\caption{Mean Signal Latency (ms)}
\label{tab:latency}
\small
\begin{tabular}{lrrrc}
\toprule
Signal & Qwen & Llama & Mistral & Pre-gen? \\
\midrule
RouteLLM & $<$1 & $<$1 & $<$1 & $\checkmark$ \\
KS (FAISS) & 72 & 72 & 77 & $\checkmark$ \\
GSA v3 & 461 & 777 & 767 & $\checkmark$ \\
logprob & 1,531 & 4,756 & 3,873 & $\times$ \\
SC & 539 & 2,874 & 2,766 & $\times$ \\
\bottomrule
\end{tabular}
\end{table}

\subsection{Naive Fusion Hurts}

Mean fusion drags the strong signal toward weaker ones (Table~\ref{tab:fusion}). The effect is largest on Qwen where the quality gap is widest (logprob 0.714 vs.\ next-best GSA 0.562). Fusion only approaches logprob on Llama, where multiple signals have similar strength.

\begin{table}[t]
\centering
\caption{AUROC of Signal Combinations (Simple Mean) vs.\ Logprob Alone}
\label{tab:fusion}
\begin{tabular}{lrrr}
\toprule
Combination & Qwen & Llama & Mistral \\
\midrule
logprob alone & \textbf{0.714} & \textbf{0.650} & \textbf{0.678} \\
all signals (mean) & 0.523 & 0.643 & 0.643 \\
logprob + GSA & 0.634 & 0.645 & 0.669 \\
\bottomrule
\end{tabular}
\end{table}

\subsection{Retrieval-Conditional Self-Assessment}
\label{sec:retrieval-gsa}

Bare self-assessment asks the model ``are you confident?'' with no external context. An intuitive improvement is to inject retrieved knowledge so the model can ground its judgment. However, naive retrieval injection \emph{hurts}: when the model sees ``no relevant knowledge retrieved'' or marginal hits, it is primed toward NO even when it could answer correctly from its own weights (a failure mode we documented during development).

We propose retrieval-conditional self-assessment (GSA~v3): inject retrieved hits \emph{only} when their similarity score exceeds a threshold $\tau$, and otherwise fall back to a prompt byte-identical to the bare variant. The model cannot distinguish ``retrieval found nothing strong'' from ``retrieval was never attempted,'' eliminating the negative priming effect.

Strong-hit-only retrieval ($\tau{=}0.70$) improves the signal by $+$0.037 on Qwen and up to $+$0.069 on Llama (Table~\ref{tab:gsa}); including marginal hits ($\tau{=}0.60$) degrades it below the bare baseline. The threshold was selected based on qualitative analysis during development: marginal retrieved hits (scores 0.50--0.65) confused the model's self-assessment by providing tangentially related context, while strong hits (${\geq}0.70$) consistently improved it. A finer-grained sweep is left to future work. The design principle---never reveal retrieval absence---is model-agnostic but density-dependent: it requires sufficient KB coverage to activate on a meaningful fraction of queries.

As a pre-generation signal (${\sim}$500ms), GSA~v3 is 3--10$\times$ cheaper than log-probability and can serve as a first-stage filter in a two-stage routing system, escalating low-confidence queries before committing to full local generation. On TriviaQA, GSA~v3 shows no improvement over bare GSA because the 200-entry KB is too sparse: at $\tau{=}0.70$, only 2.6\% of queries triggered the retrieval prompt, confirming the density dependence.

\begin{table}[t]
\centering
\caption{GSA v3 (Retrieval-Conditional) vs.\ v2 (Bare) on Qwen MMLU-Pro ($n{=}931$). Threshold $\tau$ controls which retrieved hits are shown.}
\label{tab:gsa}
\begin{tabular}{lrr}
\toprule
Variant & AUROC & Queries w/ Retrieval \\
\midrule
GSA v2 (bare) & 0.562 & 0 / 931 \\
GSA v3 ($\tau{=}0.70$) & \textbf{0.599} & 290 / 931 (31\%) \\
GSA v3 ($\tau{=}0.60$) & 0.511 & 769 / 931 (83\%) \\
\bottomrule
\end{tabular}
\end{table}

\section{Analysis}
\label{sec:analysis}

\subsection{Why Does Log-Probability Work?}
\label{sec:analysis-logprob}

Average token log-probability measures how surprised the model was by its own output. On factual question answering (QA), a model that ``knows'' the answer produces high-probability tokens; a model that is guessing produces lower-probability, higher-variance tokens~\cite{kadavath2022language}.

Across our three model families, the signal's strength correlates with RLHF training emphasis (Table~\ref{tab:rlhf}). Models with calibration-aware RLHF produce stronger logprob signals; models trained primarily for helpfulness benefit more from explicit self-assessment. Three models cannot establish a causal relationship, but the pattern is consistent with the hypothesis that calibration-focused training preserves token-level probability information. If this holds more broadly, it suggests the two signals are complementary across the model-family axis---a testable prediction for future work.

\begin{table}[t]
\centering
\caption{Signal Quality vs.\ RLHF Emphasis}
\label{tab:rlhf}
\begin{tabular}{llrr}
\toprule
Model & RLHF Emphasis & logprob & GSA \\
\midrule
Qwen-2.5-7B & Calibration-aware & \textbf{0.714} & 0.562 \\
Mistral-7B & Mixed & 0.678 & 0.638 \\
Llama-3.1-8B & Helpfulness & 0.650 & \textbf{0.614} \\
\bottomrule
\end{tabular}
\end{table}

\subsection{Why Does Supervised Routing Collapse Cross-Dataset?}

RouteLLM learns which regions of embedding space correspond to easy queries on MMLU-Pro. This transfers across model families (Qwen-trained RouteLLM achieves 0.663--0.675 on Llama/Mistral) because query difficulty is partially model-agnostic. But the embedding structure of MMLU-Pro is specific to MMLU-Pro. TriviaQA occupies a different region with different difficulty patterns.

Log-probability measures a property of the model's generation, not of the query. It transfers because it does not depend on the query distribution.

\subsection{Why Does Fusion Hurt?}

Equal-weight fusion drags a strong signal toward weaker ones---a known failure mode when component quality is heterogeneous~\cite{dietterich2000ensemble}. Thompson Sampling fusion, which should learn to upweight strong signals, collapses to equal-weight in our setup because no routing decisions are made during evaluation. Adaptive fusion with real feedback remains open.

\subsection{Knowledge Similarity: A Same-Dataset Confound}

KS is consistently inverted (AUROC 0.413--0.426). Per-category analysis reveals the mechanism: harder MMLU-Pro categories (law at 16.2\% accuracy, engineering at 29.2\%) have \emph{higher} retrieval similarity because the KB was seeded from the same distribution. A KB seeded from external sources would produce a different distribution. We did not test this.

\section{Cold-Start Deployment Guide}
\label{sec:deployment}

Table~\ref{tab:deploy} summarizes routing options at deployment time.

\textbf{Recommendation.} Use logprob from the first query. If latency is critical, add GSA as a pre-generation filter. RouteLLM training is only justified when (a)~the query distribution is known and stable, (b)~pre-generation routing is required, and (c)~cross-dataset transfer is not needed. For factual QA deployments at the 7--8B scale, logprob at \$0 is the better default.

\begin{table}[t]
\centering
\caption{Routing Options at Deployment Time}
\label{tab:deploy}
\small
\begin{tabular}{llccc}
\toprule
Stage & Signal & AUROC & Cost & Latency \\
\midrule
Day 0 & logprob & .650--.833 & \$0 & 1.5--4.8s \\
Day 0 & GSA v3 & .562--.720 & \$0 & 0.5--0.8s \\
After label. & RouteLLM & .620--.676 & ${\sim}$\$25 & $<$1ms \\
\bottomrule
\end{tabular}
\end{table}

\textbf{Operating points.} Table~\ref{tab:operating} shows the cost--quality tradeoff when routing with logprob on MMLU-Pro, averaged across three models. Queries are ranked by logprob confidence; the bottom $k$\% are escalated to cloud. At 80\% escalation, 94\% of cloud quality is preserved while eliminating 18\% of cloud cost. The tradeoff steepens with lower local accuracy: Qwen (43.7\% local) reaches 97\% quality preservation at 80\% escalation, while Mistral (30.5\% local) reaches 92\%. On tasks where local accuracy is higher (e.g., TriviaQA at 55--72\%), the savings at equivalent quality would be substantially larger.

\begin{table}[t]
\centering
\caption{Logprob Routing Operating Points on MMLU-Pro (Averaged Across 3 Models). Bottom $k$\% by Logprob Escalated to Cloud.}
\label{tab:operating}
\small
\begin{tabular}{rrrr}
\toprule
Escalate \% & Local \% & Accuracy & Quality Pres. \\
\midrule
0 (local only) & 100 & 35.8\% & 0.46 \\
20 & 80 & 45.3\% & 0.58 \\
40 & 60 & 55.8\% & 0.71 \\
60 & 40 & 65.1\% & 0.83 \\
80 & 20 & 74.1\% & 0.94 \\
100 (cloud only) & 0 & 78.7\% & 1.00 \\
\bottomrule
\end{tabular}
\end{table}

\section{Related Work}
\label{sec:related}

\textbf{Supervised routing.} RouteLLM~\cite{ong2024routellm} established the dominant paradigm: train a classifier on labeled (query, model-was-sufficient) pairs. Subsequent work extends this to contextual bandits under budget constraints~\cite{chen2025adaptive}, bandit feedback with regret guarantees~\cite{soare2025bandit}, dueling feedback~\cite{li2025dueling}, non-stationary pricing~\cite{tabernermiller2026paretobandit}, and routing strategy surveys~\cite{gomes2025routing}. All require training data---labeled examples, preference pairs, or online feedback. In our experiments, RouteLLM training costs approximately \$25 and 2~hours per local model. None of these works benchmark against zero-shot alternatives.

\textbf{Uncertainty-based routing.} Chuang et al.~\cite{chuang2025confident} benchmark uncertainty-based SLM routing across 1,500+ settings. Zhang et al.~\cite{zhang2025uncertainty} propose confidence-driven edge-cloud routing. Mukkunnoth et al.~\cite{mukkunnoth2025confidence} combine three signals for pre-generation hallucination routing. Our work differs in directly comparing against supervised baselines and testing cross-dataset transfer.

\textbf{Activation probes.} Most directly related, Lugoloobi et al.~\cite{lugoloobi2026encode} train linear probes on pre-generation activations to predict model success, achieving AUROC ${>}0.7$ on math and coding tasks with up to 70\% cost reduction. They find routing effectiveness is limited by probe reliability---a conclusion we reach independently. Their approach requires model internals (residual-stream activations) and per-model training. Our logprob signal requires only token probabilities from any API, works zero-shot, and transfers across datasets.

\textbf{Token-level calibration.} Kadavath et al.~\cite{kadavath2022language} showed calibrated token probabilities in RLHF'd models. Guo et al.~\cite{guo2017calibration} established calibration theory for deep networks. Huang et al.~\cite{huang2024logprob} confirmed log-probability reliability in instruction-tuned models. We apply the same mechanism to routing and show it transfers where supervised approaches fail.

\textbf{Self-assessment.} Kadavath et al.~\cite{kadavath2022language} tested P(True). Ren et al.~\cite{ren2023selfeval} reformulated generation into self-evaluation. Chen et al.~\cite{chen2024trust} found self-reported confidence outperforms self-consistency at lower cost. All use fixed prompts without external context. We extend self-assessment with retrieval-conditional prompting---selectively injecting retrieved knowledge when similarity is high and falling back to an indistinguishable bare prompt otherwise---and show that this design improves signal quality while naive always-inject or always-bare approaches do not.

\textbf{Multi-signal uncertainty.} UQLM~\cite{bouchard2026uqlm,bouchard2025uncertainty} and MUSE~\cite{geng2025muse} combine signals into unified frameworks. We find naive fusion hurts---consistent with ensemble theory~\cite{dietterich2000ensemble} under heterogeneous quality.

\textbf{Self-consistency.} Wang et al.~\cite{wang2022selfconsistency} proposed multi-path agreement. Our two-sample variant is model-specific: effective on Llama/Mistral, dead on Qwen.

\textbf{Semantic entropy.} Kuhn et al.~\cite{kuhn2023semantic} proposed detecting confabulations via entropy over semantic clusters. Farquhar et al.~\cite{farquhar2024robust} extended this with cheaper approximations. These methods require 5--10 generations per query, making them 5--10$\times$ more expensive than our single-generation logprob signal.

\textbf{Verbalized confidence.} Tian et al.~\cite{tian2023just} showed LLMs can express calibrated confidence when prompted. Xiong et al.~\cite{xiong2024llmsexpress} found verbalized confidence is often overconfident. Our GSA signal extracts YES-token probability rather than a numeric score, avoiding the overconfidence problem.

\textbf{Confidence biases.} Kumaran et al.~\cite{kumaran2026competing} showed that LLMs exhibit two competing biases: a choice-supportive bias (overconfidence when the model sees its own prior answer) and overweighting of opposing advice (underconfidence when presented with contradictory information). These findings provide mechanistic support for our observation that generation-side signals like logprob---which bypass explicit metacognitive processes---outperform self-assessment signals that are subject to such biases.

\textbf{Surveys.} Geng et al.~\cite{geng2024survey} and Huang et al.~\cite{huang2025uncertainty} survey LLM uncertainty quantification. Our work is narrower but deeper: fewer methods, more models and datasets, with a supervised baseline comparison.

\section{Limitations}
\label{sec:limitations}

\begin{enumerate}
\item \textbf{Two datasets, one task type.} Both MMLU-Pro and TriviaQA are factual QA. Reasoning, creative, and multi-turn tasks are untested.
\item \textbf{Three models, one size.} All 7--8B. The calibration hypothesis predicts logprob improves with scale; untested.
\item \textbf{Post-generation routing cost.} Logprob requires full generation (1.5--4.8s) before routing. Wasted on escalated queries.
\item \textbf{Static KB.} All experiments use a frozen knowledge base. In production, the KB grows from escalations, making routing non-stationary.
\item \textbf{MCQ inflation.} MMLU-Pro's 10-option format gives a 10\% chance floor. TriviaQA partially addresses this.
\item \textbf{Label noise.} 13\% inter-judge disagreement. Absolute AUROC carries ${\sim}{\pm}0.05$ uncertainty; differential comparisons are robust.
\end{enumerate}

\section{Conclusion}
\label{sec:conclusion}

Supervised routing learns properties of the query distribution; zero-shot confidence signals measure properties of the model's generation. This distinction explains our central finding: average token log-probability---available from the first query at zero cost---matches supervised baselines in-distribution (AUROC 0.650--0.714 vs.\ 0.644--0.676) and substantially outperforms them when the query distribution shifts (0.717--0.833 vs.\ 0.512--0.564), because it does not depend on the input distribution at all. A supervised baseline trained on 1,000 labeled examples never exceeds the zero-shot signal, and naive multi-signal fusion hurts rather than helps.

For practitioners deploying local-to-cloud routing on factual QA tasks: start with log-probability. When pre-generation latency matters, retrieval-conditional self-assessment---which selectively grounds the model's confidence judgment in retrieved knowledge---provides a cheaper first-stage filter at 3--10$\times$ lower latency. The supervised training investment is hard to justify when the zero-shot alternative matches it in-distribution and exceeds it out-of-distribution.

The main limitation is scope: two factual QA datasets and three 7--8B models. Whether the generation-side advantage holds on reasoning tasks, creative tasks, or larger models remains open. The mechanism---token-level surprise---is task-agnostic in principle, but its effectiveness likely varies: a model can produce high-probability tokens for a confidently wrong answer on tasks where surface fluency does not correlate with correctness. The evidence for broader generalization is not yet in hand.

\bibliographystyle{IEEEtran}

\end{document}